\documentclass[sigconf]{acmart}
\AtBeginDocument{%
  }

\copyrightyear{2025}
\acmYear{2025}
\setcopyright{cc}
\setcctype{by}
\acmConference[MM '25]{Proceedings of the 33rd ACM International Conference on Multimedia}{October 27--31, 2025}{Dublin, Ireland}
\acmBooktitle{Proceedings of the 33rd ACM International Conference on Multimedia (MM '25), October 27--31, 2025, Dublin, Ireland}
\acmDOI{10.1145/3746027.3755821}
\acmISBN{979-8-4007-2035-2/2025/10}

\usepackage{multirow}
\usepackage{soul}
\usepackage{caption}
\usepackage{newfloat}
\DeclareFloatingEnvironment[placement={!ht},name=List]{mylist} 
\begin{document}

\title{SD-VSum: A Method and Dataset for Script-Driven Video Summarization}

\author{Manolis Mylonas}
\affiliation{%
  \institution{ITI, CERTH}
  \city{Thessaloniki}
  \country{Greece}}
\email{emylonas@iti.gr}

\author{Evlampios Apostolidis}
\affiliation{%
  \institution{ITI, CERTH}
  \city{Thessaloniki}
  \country{Greece}}
\email{apostolid@iti.gr}

\author{Vasileios Mezaris}
\affiliation{%
  \institution{ITI, CERTH}
  \city{Thessaloniki}
  \country{Greece}}
\email{bmezaris@iti.gr}

\renewcommand{\shortauthors}{Manolis Mylonas, Evlampios Apostolidis, and Vasileios Mezaris}

\begin{abstract}
In this work, we introduce the task of script-driven video summarization, which aims to produce a summary of the full-length video by selecting the parts that are most relevant to a user-provided script outlining the visual content of the desired summary. Following, we extend a recently-introduced large-scale dataset for generic video summarization (VideoXum) by producing natural language descriptions of the different human-annotated summaries that are available per video. In this way we make it compatible with the introduced task, since the available triplets of ``video, summary and summary description'' can be used for training a method that is able to produce different summaries for a given video, driven by the provided script about the content of each summary. Finally, we develop a new network architecture for script-driven video summarization (SD-VSum), that employs a cross-modal attention mechanism for aligning and fusing information from the visual and text modalities. Our experimental evaluations demonstrate the advanced performance of SD-VSum against SOTA approaches for query-driven and generic (unimodal and multimodal) summarization from the literature, and document its capacity to produce video summaries that are adapted to each user's needs about their content.
\end{abstract}





\maketitle

\section{Introduction}
\begin{figure*}[t]
\centering
\includegraphics[width=0.95\textwidth]{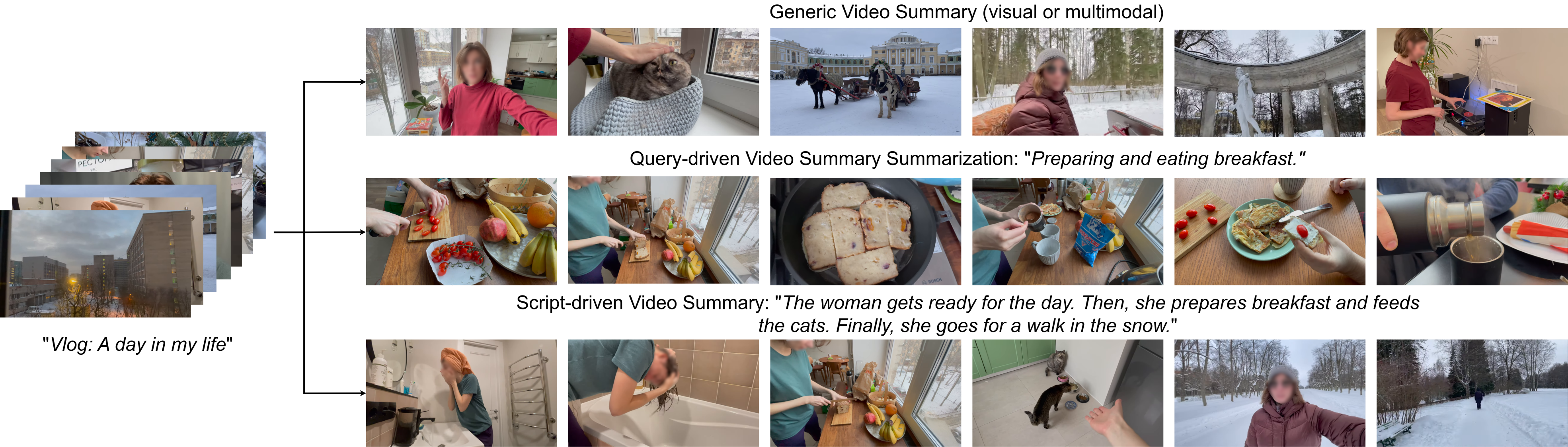}
\caption{A video blog (vlog) with the daily activities of the vlogger, and the produced summaries by: generic (top row), query-driven (middle row), and script-driven video summarization (bottom row). The generic video summary aims to provide a complete overview of the video. The query-driven video summary concentrates solely on parts of the video that match with a short user query. The script-driven video summary offers a more tailored synopsis of the video, containing parts that relate to the different activities outlined in the user script and avoiding parts that are less interesting for the user, and thus redundant.}
\label{fig:task}
\end{figure*}

The established approach in the Media industry for producing summarized versions of a full-length video for distribution via different communication channels (e.g., social networks and video sharing platforms), requires the visual inspection of the entire content by a professional editor who then decides about the parts that should be included in the summary. This time-demanding and laborious task can be facilitated by modern AI-based video summarization technologies \cite{9594911}. Nevertheless, these technologies can assist the work of editors only to a certain extent, since they typically learn how to create video summaries based on generic summarization criteria, such as the representativeness and diversity of the summary (see the top part of Fig. \ref{fig:task}), and thus, are not able to generate video summaries according to the editor's needs about their content.

Being more tailored to the aforementioned requirement, a number of text/query-driven video summarization methods have appeared \cite{10.1007/978-3-319-46484-8_1,8099712,10.5555/3504035.3504062,10.1145/3372278.3390695,10.5555/3540261.3541333,10.1007/978-3-031-19830-4_31,10204014}. These methods allow to generate video summaries that are more customized to the users' preferences. However, these preferences are typically expressed using one or more keywords (e.g., ``food'', ``cooking pancakes'') \cite{10.1007/978-3-319-46484-8_1,8099712,10.5555/3540261.3541333} or a short sentence (e.g., ``a woman is making a sandwich'') \cite{10.5555/3504035.3504062,10.1145/3372278.3390695}; as a consequence, the generated summaries by these methods show limited visual and semantic diversity, as they mainly contain parts of the video that match such a short user query (see the middle part of Fig. \ref{fig:task}). The existing text/query-driven video summarization methods are not compatible with more extensive descriptions, formed by multiple sentences that can relate to different events depicted in the video, and possibly include abstract cues and narrative elements. This is a different problem that necessitates a different technical approach for solving it.

Based on the above, in this paper we introduce the task of script-driven video summarization, whose goal is to generate a summary of the full-length video by choosing the parts which are most relevant to a longer user-provided query (``script'') that outlines the content of the desired summary. To support research on this task, we extend a recently-introduced large-scale dataset for video summarization (VideoXum \cite{10334011}), by producing natural language descriptions of the visual content of the different human-annotated summaries that are available per video, using the SOTA Large Multimodal Model LLaVA-NeXT-Video-7B \cite{li2024llavanext-strong}. The multiple triplets of ``video, summary and summary description'' that are available for each video in the extended S-VideoXum dataset can be used to train a method to produce different summaries for a given video, based on the provided description (the so-called script) about the content of each summary. Finally, we build a new method for script-driven video summarization (called SD-VSum), that employs a cross-modal attention mechanism to align and fuse information from the visual and text modalities, and computes the frames' importance using a trainable Transformer-based scoring mechanism. SD-VSum is designed to generate video summaries that are visually and semantically more diverse (see the bottom part of Fig. \ref{fig:task}) compared to summaries created by text/query-driven summarization methods, and less redundant compared to summaries produced by generic summarization approaches. We train and evaluate SD-VSum using S-VideoXum and compare its performance against SOTA approaches for query-driven and generic (unimodal and multimodal) video summarization from the literature. Our quantitative and qualitative evaluations highlight the advanced performance of SD-VSum and demonstrate its capacity to generate video summaries that meet the users' needs. Our contributions are as follows:
\begin{itemize}
    \item We introduce the task of script-driven video summarization, which aims to produce a summary of the full-length video by selecting the parts that are most relevant to a user-provided script about the visual content of the summary.
    \item We extend the recently-introduced VideoXum large-scale dataset for video summarization to S-VideoXum, by producing natural language descriptions of the different human-annotated summaries that are available per video.
    \item We propose a new method for script-driven video summarization (SD-VSum) that employs a cross-modal attention mechanism; we use the extended S-VideoXum dataset to train and evaluate this method, and document its competitiveness against other SOTA approaches for query-driven and generic video summarization.
\end{itemize}

\section{Related Work}

\subsection{Text/Query-driven video summarization}

In one of the first attempts towards text-driven video summarization, Sharghi et al. (2016) \cite{10.1007/978-3-319-46484-8_1} proposed a probabilistic model for query-focused video summarization, which selects the video shots that jointly have the highest relevance to the user query and the highest importance in the context of the video. To avoid the use of costly supervision data, in their subsequent work, Sharghi et al. (2017) \cite{8099712} described a memory network parameterized probabilistic model that implicitly attends the user query about the video onto different video frames and shots. Vasudevan et al. (2017) \cite{10.1145/3123266.3123297} employed a submodular optimization framework to form a query-adaptive video summarization method that selects the frames of the video summary by quantifying their relevance with the text query based on the distance of deep representations in a common embedding space, and taking into account frames' diversity, representativeness and aesthetic quality. Wei et al. (2018) \cite{10.5555/3504035.3504062} presented a semantic attended video summarization network that contains a frame selector and a video descriptor, and learns how to choose the most semantically representative video parts by minimizing the distance between the generated description for the video summary and human-provided descriptions of visually-coherent segments of the video. Zhang et al. (2018) \cite{Zhang2018QCVS} used a three-player loss to train a query-conditioned generative adversarial network, where the generator creates a summary based on the learned joint representation of the user query and the video content, and the discriminator tries to discriminate the real summary from a generated and a random one. Jiang et al. (2019) \cite{10.1145/3323873.3325040} designed a hierarchical network architecture to model query-related long-range temporal dependency, and combined diverse attention mechanisms used to encode query-related and context-important information, with a multi-level self-attention and a variational autoencoder used to incorporate user-oriented diversity and stochastic factors. Huang et al. (2020) \cite{10.1145/3372278.3390695} described a method for query-controllable video summarization, which uses: a summary controller that describes the desired video summary, a summary generator that scores frames based on their relationship with the textual description, and a summary output module that creates a summary by choosing the top-k scoring frames. Xiao et al. (2020) \cite{Xiao_Zhao_Zhang_Yan_Yang_2020} modeled the high-level semantics of each video shot using local self-attention mechanisms and the semantic relationship between all shots and a given query using a query-aware global-attention, and employed a query-relevance ranking module to compute a similarity score for each shot and a given query, and form the summary. To eliminate the bias when the model is trained on a small dataset, in their next work, Xiao et al. (2020) \cite{9063637} trained a hierarchical self-attentive network for estimating the importance of video frames and shots using the ActivityNet Captions dataset \cite{8237345}, fine-tuned it using a reinforced caption generator that constructs a video description using the selected frames, and employed a query-aware scoring module that computes shot-level scores for a given query and produces the video summary. Cizmeciler et al. (2021) \cite{10.1007/s11042-022-12442-w} presented a query-focused video summarization method that employs semantic attributes as indicators of query relevance and semantic saliency maps from pretrained attribute/action classifiers to locate relevant regions in the video frames. Narasimhan et al. (2021) \cite{10.5555/3540261.3541333} introduced the CLIP-It method for query-focused video summarization, which uses a multi-head language-guided attention mechanism that learns how to estimate the frames importance based on their visual relevance and their correlation with a user-defined query. Formulating the task of query-attentive video summarization as a multi-label classification problem, Hu et al. (2023) \cite{Hu2023QuerybasedVS} predicted the correlation between visual content and multi-concept labels by feeding deep features into a multi-layer perceptron, and selected the video part with the highest relevance to the user's query to form the summary. Finally, to deal with the lack of large labeled datasets for query-based video summarization, Huang et al. (2023) \cite{10222138} resorted to the use of self-supervision and the generation of segment-level pseudo labels from input videos to model the relationship between a pretext and a target task and between the pseudo label and the human-defined label, and employed a semantics booster to generate context-aware query representations and mutual attention to capture the interaction between visual and textual modalities. For further details, we refer the reader to a recent survey on query-attentive video summarization \cite{Kadam2024}.

The methods described above create a video summary taking into account the user's preferences about its content, expressed using a few keywords \cite{10.1007/978-3-319-46484-8_1,8099712,10.1145/3123266.3123297,Zhang2018QCVS,10.1145/3323873.3325040,10.1145/3372278.3390695,Xiao_Zhao_Zhang_Yan_Yang_2020,10.1007/s11042-022-12442-w,9063637,10.5555/3540261.3541333,Hu2023QuerybasedVS,10222138} or a short sentence \cite{10.5555/3504035.3504062,10.5555/3540261.3541333}. Contrary to these methods, SD-VSum guides the summarization process using a free-form textual description of the content of the desired video summary, allowing the user to create summaries that are visually and semantically more diverse compared to the summaries generated by the existing text/query-driven summarization methods.

\subsection{Multimodal video summarization}

Closely related to the text/query-driven videos summarization methods, in terms of the used data modalities, are multimodal approaches for generic video summarization which take as input the full-length video and a textual description of its content and/or the video transcripts. The CLIP-It method of Narasimhan et al. (2021) \cite{10.5555/3540261.3541333} for query-based video summarization is also capable to produce generic video summaries using automatically-generated dense captions of the video content. In their subsequent work, Narasimhan et al. (2022) \cite{10.1007/978-3-031-19830-4_31} described a transcript-based method for summarizing instructional videos by selecting important steps of the procedure that are most relevant to the task, as well as salient in the video, i.e., referenced in the speech. Zhong et al. (2022) \cite{10.1145/3477538} presented an unsupervised video summarization approach that learns how to estimate the frames' importance and create semantically representative video summaries, by minimizing the distance of learnable representations of the video content and its textual description in a common latent space. He et al. (2023) \cite{10204014} presented a unified multimodal transformer-based model which can align and attend multimodal inputs leveraging time correspondence (i.e., video and transcript), and selects the most important frames of the video (to form a keyframe-based video summary) and the most importance sentences of the transcript (to produce a text summary). Finally, Argaw et al. (2024) \cite{10656029} trained a multimodal method for video summarization using ground-truth pseudo-summaries obtained by prompting a Large Language Model to extract the most critical and informative moments from ASR transcripts and mapping back the resulting textual summary to the relevant video segments. Their method creates a video summary by taking into account the visual content and a long-form description of it, or the video transcripts.

The aforementioned multimodal approaches utilize additional information about the content of the full-length video. As a consequence, the generated summaries provide a synopsis of the entire video and they are not adapted to any specific requests about their content. Differently from these methods, SD-VSum takes into account information about the content of the desired summary, and thus it is capable to produce video summaries that do not contain redundant information and are customized to the users' demands.

%\vspace{-1mm}
\subsection{Datasets}

\begin{table*}[t]
\caption{Overview of the different datasets for text/query-driven video summarization from the literature. The original VideoXum dataset that is used as a basis for our S-VideoXum dataset, is included as a reference.}
\label{tab:datasets}
\begin{tabular}{|l|c|c|c|c|c|c|}
\hline
Dataset  & Domains & Videos & Duration            & \begin{tabular}[c]{@{}c@{}}Annotations\\ per video\end{tabular} & \begin{tabular}[c]{@{}c@{}}Type of annotations for text/\\ query-driven summarization \end{tabular} & Data split                   \\  \hline
UTE \cite{6247820} & 1             & 4         & 3 - 5 hr.           & 3      & textual summaries                                                         & train, test   \\
TVE \cite{Yeung2014VideoSETVS} & 1         & 4    & 45 min.          & 3   & textual summaries                                                            & train, test     \\
QFVS \cite{8099712}    & 1             & 4         & 3 - 5 hr.           & 46  & query-based video summaries                                                            & train, test   \\
RAD \cite{10.1145/3123266.3123297}    & 22            & 190        & 2 - 3 min.         & 5 & query-relevance labels                                                             & train, validation, test  \\
SumMe \cite{10.1007/978-3-319-10584-0_33}     & -             & 25        & 1 - 6 min.          & 3-5  & fragment-level descriptions                                                         & train, test   \\
TVSum \cite{7299154}   & 10            & 50        & 1 - 11 min.         & 3-5    & fragment-level descriptions                                                          & train, test     \\
ARS \cite{10.1007/s11042-022-12442-w} & 2   & 10    & 2 - 12 min. & 10    & query-based video summaries                                                          & train, test    \\ \hline
VideoXum \cite{10334011} & open domain   & 14,000    & up to 12.5 min. & 10   & -                                                           & train, validation, test  \\
S-VideoXum (Ours) & open domain   & 11,908    & up to 12.5 min. & 10    & video summary descriptions                                                          & train, validation, test  \\ \hline
\end{tabular}
\end{table*}

An overview of the datasets from the literature for text/query-driven video summarization is given in Table \ref{tab:datasets}. The earliest used ones by Sharghi et al. in \cite{10.1007/978-3-319-46484-8_1}, were UT Egocentric (UTE) \cite{6247820} and TV Episodes (TVE) \cite{Yeung2014VideoSETVS}. UTE contains four egocentric videos $3-5$ hours long that were recorded in an uncontrolled environment from a first-person view wearable camera. TVE includes four videos $45$ min. long that were captured from a third person's viewpoint. The videos of the UTE and TVE have been annotated by human subjects in \cite{Yeung2014VideoSETVS} with one-sentence descriptions at the level of short video fragments that last $5$ and $10$ sec., respectively. Three reference text-based summaries were defined per video as a subset of the textual annotations, and used for evaluating query-driven summarization based on the ROUGE-SU metric \cite{lin-2004-rouge}. UTE was further extended by Sharghi et al. in \cite{8099712}, who added human-annotated video summaries for $46$ user queries that were defined using two or three concepts per video, forming the Query-Focused Video Summarization (QFVS) dataset. Vasudevan et al. \cite{10.1145/3123266.3123297} introduced the Relevance and Diversity (RAD) dataset which contains $190$ videos that have been retrieved from YouTube using different queries and annotated with query relevance labels at the frame and fragment level, ranging from 0 (Bad) to 3 (Very Good). This dataset was extended by Huang et al. in \cite{10.1145/3372278.3390695}, who computed relevance scores for each query-video pair of the dataset. The SumMe \cite{10.1007/978-3-319-10584-0_33} and TVSum \cite{7299154} benchmarking datasets for generic video summarization were extended by Wei et al. \cite{10.5555/3504035.3504062} with fragment-level descriptions of the visual content, to train a video summarization method that takes into account the video semantics. Finally, Cizmeciler et al. \cite{10.1007/s11042-022-12442-w} introduced the Activity Related Summaries (ARS) dataset containing $10$ YouTube videos ($2$-$12$ min. long) that have been annotated by humans at the shot level using a list of $60$ attributes, while $10$ ground-truth summaries were defined per video using a list of $142$ one/two-word queries.

As shown in Table \ref{tab:datasets}, contrary to the existing datasets, the extended S-VideoXum dataset for script-driven video summarization: i) contains a remarkably large number of videos from different domains with diverse visual content, ii) provides $10$ different ground-truth summaries per video, thus allowing to train a summarization method to produce different summaries for a given video based on the users' preferences about the content of each summary, and iii) is divided into train, validation and test sets, enabling the selection of a well-trained model that generalizes well to unseen data. 

\section{Problem Statement and Dataset}

\subsection{Problem formulation}

Let us assume an input video of $N$ frames and a user-provided script (a free-form textual description) of $M$ sentences outlining the content of the desired video summary; different sentences of the script may refer to different parts of the full-length video with varying visual and semantic content. Then, script-driven video summarization aims to select a subset of frames or fragments of the video that are both relevant to one or more sentences of the user-provided script, and important for an overall understanding of the video. The selected frames/fragments should form a video summary that does not exceed a predefined length $L$, which is typically set to $15\%$ of the full-length video duration \cite{9594911}. 

Another task that aims to spot parts of the video that are relevant to a freeform query is the task of temporal sentence grounding in videos \cite{10.1145/3532626} (a.k.a. video grounding). Nevertheless, this task tries to localize one target segment from the full-length video, that corresponds to a given sentence query. On the contrary, script-driven video summarization aims to generate a concise summary, and in this process it needs to select multiple video fragments that relate to the complex query, as well as other less relevant parts that are necessary for providing a synopsis of the full-length video.
 
\subsection{Script-driven video summarization dataset}
\label{subsection:sdvs}

\begin{figure*}[t]
\centering
\includegraphics[width=0.99\textwidth]{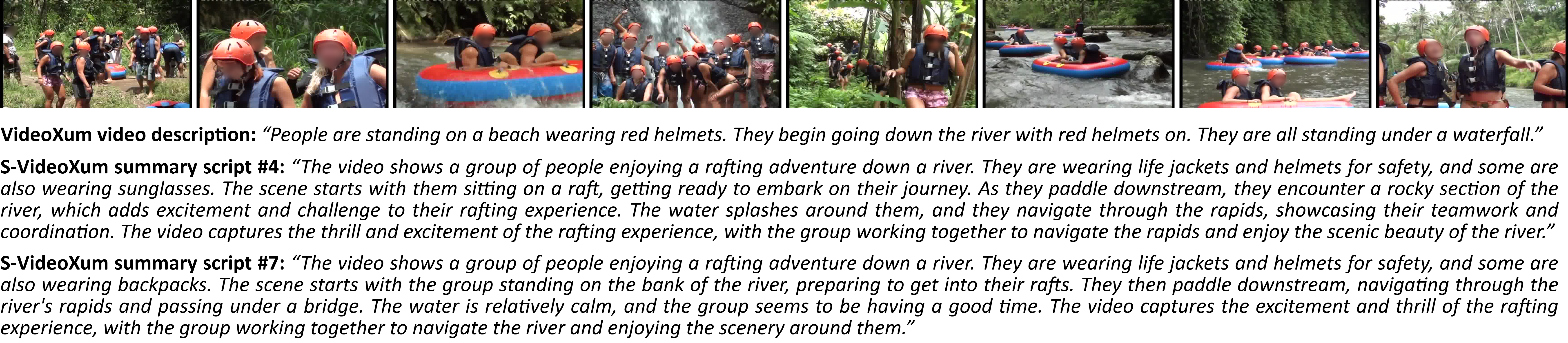}
\caption{A keyframe-based representation of the video ``v\_3OcAjx8e4LU'', the textual annotation in the original VideoXum dataset, and examples of the generated scripts for two ground-truth summaries of this video in the S-VideoXum dataset.}
\label{fig:annot-examples}
\end{figure*}

To overcome the observed weaknesses of small video summarization datasets, related to the sensitivity of the evaluation outcomes to different data splits \cite{10.1145/3394171.3413632} and the inability to select a well-trained model in the absence of a validation set \cite{9594911}, we focused on the VideoXum large-scale dataset for cross-modal video summarization \cite{10334011}. VideoXum includes $14,000$ open-domain videos up to $12.5$ min. long with diverse visual content, from the ActivityNet Captions dataset \cite{8237345}. Each video is associated with multiple ground-truth video summaries - in the form of frame-level binary scores which denote the inclusion or not of a frame in the video summary - that were obtained by $40$ different human annotators. The dense video captions from ActivityNet Captions, that are considered as text summaries in \cite{10334011}, provide a high-level description of the full-length video (see Fig. \ref{fig:annot-examples}) and thus are more appropriate for multimodal generic video summarization purposes. The existence of multiple ground-truth summaries per video ($10$ in total) makes VideoXum well-suited for the script-driven video summarization task, as it allows to train a summarization method to generate different summaries for a given video, driven by the provided script about the content of each summary. Furthermore, its significantly large size allows to split the available videos into train, validation and test set. Such a data split supports the extensive training of a model using a very wide set of training samples, the selection of a well-trained model that exhibits good generalization capacity through the use of an adequately big validation set, and the robust evaluation of its performance using a large number of testing samples.

To make VideoXum suitable for training and evaluation of script-driven video summarization methods, we extended it by producing natural language descriptions (a.k.a. scripts) of the different available human-annotated ground-truth summaries per video. For this, we employed the publicly-available\footnote{https://huggingface.co/llava-hf/LLaVA-NeXT-Video-7B-hf} SOTA Large Multimodal Model LLaVA-NeXT-Video-7B \cite{li2024llavanext-strong}. We chose this model because of its strong video understanding capabilities, outperforming other open-source models (e.g., LLaMA-VID) and performing comparably to Gemini Pro\footnote{https://llava-vl.github.io/blog/2024-04 -30-llava-next-video/}. To align with the adopted video representation strategy in \cite{10334011}, we processed each ground-truth summary using one frame per second and obtained the scripts using the following prompt: ``Describe the important scenes in the video''. The maximum number of generated tokens was set to $200$, and a $4$-bit quantization was applied to reduce computational cost. The obtained scripts are diverse. Taking into account one summary per video (the first of each video's ten summaries), the scripts vary across videos from very concise to extensive ones (\#words: min = $4$, max = $220$, avg = $126.5$, std = $30.0$; \#sentences: min = $1$, max = $32$, avg = $6.1$, std = $1.5$). To estimate the within-video diversity of summaries, we calculated the above statistics separately for each video across its ten summaries, and then averaged over the dataset. The average std of \#words and \#sentences are $21.7$ and $1.1$, i.e., within-video diversity is also quite high. Visual inspection of sample scripts verified that they are quite diverse and detailed, can refer to multiple unrelated actions, and might contain abstract cues and narrative elements (e.g., see Fig. \ref{fig:annot-examples}). In addition, the annotations in S-VideoXum correspond to individual ground-truth summaries, rather than the full-length video (as in VideoXum), and focus on different elements of the summary generated by each different annotator, as depicted in Fig. \ref{fig:annot-examples}. Finally, to allow performance comparisons with multimodal approaches for generic video summarization that use the same data modalities, we produced natural language descriptions of the full-length videos, as well. Once again, we sampled these videos keeping one frame per second and generated descriptions using LLaVA-NeXT-Video-7B and the same prompt. 

Since some of the videos from the VideoXum dataset were not publicly-available during this work, the extended S-VideoXum dataset for script-driven video summarization differs from the original one in terms of number of videos, including data for $11,908$ videos. However, a statistical analysis on the obtained videos showed that our dataset maintains the same distribution of train, validation and test samples with VideoXum. In particular, we use $6,782$ samples for training, $1,707$ for validation and $3,419$ for testing. 

Prior to constructing the S-VideoXum dataset, we experimented with a small proprietary dataset (S-NewsVSum) of a Media company, including $45$ videos from news broadcast and the associated professionally-edited video summaries and user-authored scripts.

Both datasets are shared at: https://github.com/IDT-ITI/SD-VSum

\section{Proposed Approach}

\subsection{Network architecture}
\label{subsec:network}

\begin{figure*}[t]
\centering
\includegraphics[width=0.98\textwidth]{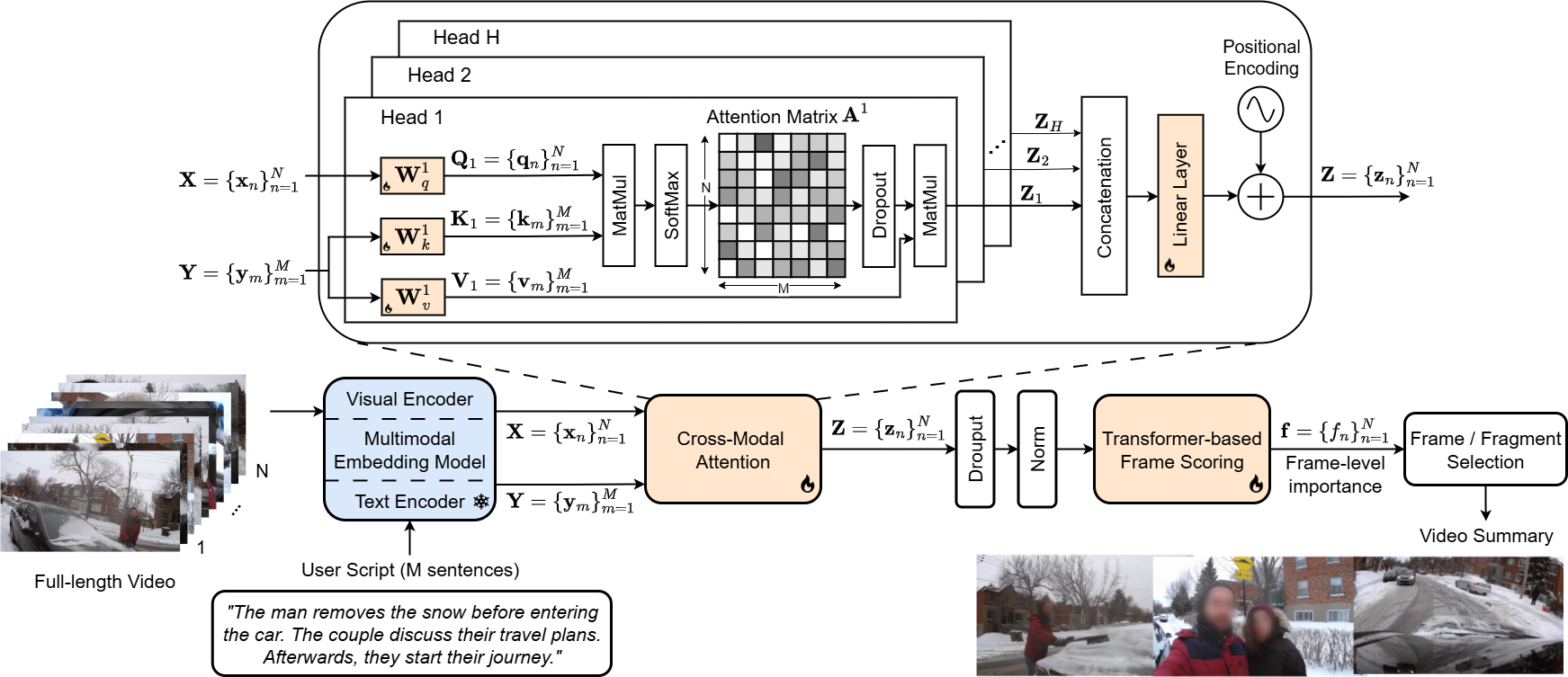}
\caption{Overview of SD-VSum. Given an input video and a user script outlining the content of the desired summary, SD-VSum generates a summary conditioned on the script. Visual and text embeddings are obtained using a pretrained multimodal embedding model, and fused using a trainable cross-modal attention mechanism, as shown in the upper part. The cross-modal embeddings in the output of this mechanism are forwarded to a trainable Transformer-based frame scoring mechanism, which computes frame-level importance scores. These scores are finally used by a frame/fragment selection component that forms the video summary based on a predefined temporal fragmentation of the video and a time-budget about the summary duration.}
\label{fig:architecture}
\end{figure*}

An overview of the SD-VSum network architecture is provided in Fig. \ref{fig:architecture}. Assuming an input video of $N$ frames (following sampling), and a user script of $M$ sentences that outlines the content of the desired summary, SD-VSum produces a video summary conditioned on the script and a time-budget about its length. A set of visual embeddings ($\boldsymbol{X} = \{\boldsymbol{x_{n}}\}_{n=1}^{N}$) of size $D$ ($\boldsymbol{x_{n}} = \{x_{r}\}_{r=1}^{D}$) are obtained for the video frames, using the visual encoder of a pretrained multimodal embedding model. Similarly, a set of text embeddings ($\boldsymbol{Y} = \{\boldsymbol{y_{m}}\}_{m=1}^{M}$) of the same size ($\boldsymbol{y_{m}} = \{y_{r}\}_{r=1}^{D}$) are computed for the different sentences of the user script (one per sentence), using the text encoder of the aforementioned model. These embeddings are then given as input to a multi-head cross-modal attention mechanism. As shown in the upper part of Fig. \ref{fig:architecture}, given the $h^{th}$ attention head of this mechanism, the visual embeddings $\boldsymbol{X}$ pass through a linear layer of size $D/H$, where $H$ denotes the number of heads, forming the Query ($\boldsymbol{Q}_{h} = \{\boldsymbol{q}_{n}\}_{n=1}^{N}$) matrix. The text embeddings $\boldsymbol{Y}$ pass through two different linear layers of size $D/H$, creating the Key ($\boldsymbol{K}_{h} = \{\boldsymbol{k}_{m}\}_{m=1}^{M}$) and Value ($\boldsymbol{V}_{h} = \{\boldsymbol{v}_{m}\}_{m=1}^{M}$) matrices. The Query and Key matrices undergo a matrix multiplication and a softmax process, formulating the values of the attention matrix ($\boldsymbol{A}_{h} = softmax(\boldsymbol{Q}_{h}\times \boldsymbol{K}_{h}^{T})\ and\ \boldsymbol{A}_{h} = \{a_{i,j}\}_{i=1,N}^{j=1,M}$). This matrix goes through a dropout layer and the outcome is multiplied with the Value matrix, defining the output of the $h^{th}$ attention head ($\boldsymbol{Z}_{h} = \{\boldsymbol{z_{n}}\}_{n=1}^{N}\ and \ \boldsymbol{z_{n}} = \{z_{r}\}_{r=1}^{D/H}$). The obtained representations in the output of all attention heads are then concatenated, and pass through a linear layer of size $D$ and a positional encoding mechanism (similar to the one in \cite{9666088}), forming the overall cross-modal embeddings ($\boldsymbol{Z} = \{\boldsymbol{z}_{n}\}_{n=1}^{N}$) in the output of the attention mechanism. Following, as shown in the lower part of Fig. \ref{fig:architecture}, these embeddings undergo a dropout and normalization process and then are given as input to a trainable Transformer-based scoring mechanism which computes frame-level importance scores ($\boldsymbol{f} = \{f_{n}\}_{n=1}^{N}$). These scores are finally used by a frame/fragment selection component that forms the video summary based on a temporal fragmentation of the video and a time-budget about the summary duration.

So, in terms of network architecture, SD-VSum is most closely related with CLIP-It \cite{10.5555/3540261.3541333}, which uses a language-guided multi-head attention mechanism to fuse information from the visual and text modalities, and employs a Transformer-based scorer for assigning importance scores to the video frames. With respect to the representation of the input data, the video is represented using similar (CLIP-based) frame-level embeddings, but the input text (query or dense video captions) is represented using a condensed single-vector embedding. The latter is produced by a trainable linear layer that gets as input a vector formed by concatenating a fixed number of sampled (CLIP-based) embeddings of the input text, and projects it into a more condensed representation (of dimension $1\times 512$). As a consequence, the language-guided attention of CLIP-It encodes the dependence between the video frames and the user query in a more compact, and thus less informative, manner. On the contrary, the cross-modal attention mechanism of the proposed SD-VSum network architecture employs a more detailed representation of the input text at the sentence level, allowing SD-VSum to encode the frames' dependence with different parts of the user script in a more effective way (using an $N\times M$ attention matrix, as shown in the upper part of Fig. \ref{fig:architecture}, rather than a more condensed $N\times 1$ matrix, as in CLIP-It). This main distinction leads to a noticeably advanced summarization performance, as documented in Section \ref{subsec:comparisons}. 

\subsection{Training regime}

Training of SD-VSum is performed in a supervised learning setting, comparing its output for a given pair of full-length video and user script, with the associated ground-truth summary (i.e., the one used to generate this particular script). Specifically, we calculate the Binary Cross Entropy (BCE) loss between the computed scores for the frames of the input video  ($\boldsymbol{f} = \{f_{n}\}_{n=1}^{N}$) by SD-VSum, and the assigned binary labels to these frames by an annotator. The calculated loss value is then back-propagated and used to update the trainable parts of the network. The use of the multiple available ground-truth summaries for each video of the dataset as individual training samples, practically increases tenfold the amount of train, validation and test samples, reported in Section \ref{subsection:sdvs}. 

A different training regime is adopted when evaluating SD-VSum on multimodal generic video summarization. In this case, the computed frame-level scores for a given pair of full-length video and textual description of its content, are compared with a single ground-truth summary that is formed by averaging the multiple human-annotated summaries for this video, at the frame-level. So, instead of comparing the output of SD-VSum with multiple different ground-truth annotations using BCE (a contradictory approach that could prevent network's training), we compute a single loss value using an average ground-truth summary and the Mean Squared Error.  

\section{Experiments}

\subsection{Evaluation protocol}

Following the evaluation approach in \cite{10334011}, we generate the video summary for a given input video and user script, by selecting the top-15\% scoring frames by a video summarization model. Then, we quantify the similarity between the machine-generated and the ground-truth summary using the F-Score (as percentage). So, a given test video is matched with each one of the multiple available user scripts for it, and each one of the generated summaries is compared with the corresponding ground-truth summary. Through this process, we compute an F-Score for each pair of compared summaries and we average these scores to form the final F-Score for this video. After performing this for all test videos, we calculate the mean of the obtained F-Score values and result in an average score that indicates the model's performance on the test set. 

When evaluating the performance on multimodal generic video summarization we also use the evaluation protocol from \cite{8954229}. In particular, we measure the alignment between the computed frame-level importance scores for a given video by a video summarization model, and the ground-truth frame-level importance scores for this video (obtained by averaging its multiple available binary ground-truth summaries, at the frame-level), using the Kendall's $\tau$~\cite{kendall1945treatment} and Spearman's $\rho$~\cite{kokoska2000crc} rank correlation coefficients. The computed $\tau$ and $\rho$ values for all test videos are then averaged, defining the performance of the video summarization model on the test set. 

The selection of a well-trained model for evaluation on the videos of the test is made based on the recorded performance on the videos of the validation set. More specifically, after each training epoch we measure the performance of the trained model on the videos of the validation set. When the training process is completed, we keep the model with the highest validation-set performance in terms of F-Score, and assess its performance on the test set using the evaluation protocols described above. 

For the small S-NewsVSum dataset, we follow the same general approach, the difference being that we have only one ground-truth summary per video. Additionally, this being a very small dataset, we perform 5-fold cross validation and report the average performance.

\subsection{Implementation details}

Similarly to \cite{10334011}, videos are represented using one frame per second, and frame-level embeddings of the visual content (of size $D=512$) are obtained using a fine-tuned CLIP model on the data of VideoXum, that has been released by the authors of \cite{10334011}\footnote{Publicly-available at: https://videoxum.github.io/}. The same model is used to extract sentence-level embeddings (of the same size $D=512$) from the generated scripts. Following the paradigm of e.g., CLIP-It \cite{10.5555/3540261.3541333}, we keep both the visual and the text encoder frozen. With respect to the SD-VSum network architecture, the number of attention heads is set equal to $8$, and the Transformer-based frame scoring mechanism consists of a Transformer network \cite{10.5555/3295222.3295349}, a linear layer with $512$ neurons, and a sigmoid function that computes the frame-level importance scores. The learning rate, dropout rate and L2 regularization factor are set equal to $5\cdot10^{-5}$, $0.5$ and $10^{-4}$, respectively. For the network's weights initialization, we use the Xavier uniform initialization approach with gain = $\sqrt{2}$ and biases = $0.1$. Training is performed in a batch mode with a batch size equal to $4$ using the Adam optimizer, and stops after $50$ epochs. All experiments were carried out on a PC with an NVIDIA RTX 3090 GPU. The PyTorch implementation of SD-VSum is publicly-available at: https://github.com/IDT-ITI/SD-VSum.

\subsection{Experimental comparisons and ablations}
\label{subsec:comparisons}

The proposed SD-VSum network architecture was compared with CLIP-It \cite{10.5555/3540261.3541333}, that exhibits SOTA performance on query-driven and multimodal (generic) video summarization. The results in the upper part of Table \ref{tab:table1} show that SD-VSum performs clearly better than CLIP-It on the extended S-VideoXum dataset, achieving an F-Score that is almost $2\%$ higher than the F-Score of CLIP-It. A similar performance gap between SD-VSum and CLIP-It was observed on the small-scale S-NewsVSum dataset; the computed F-Scores for SD-VSum and CLIP-It were $22.4$ and $20.0$, respectively.

Given the diversity of the generated scripts, we analyzed the performance of SD-VSum across varying script lengths using four subsets of S-VideoXum: S1: $10\%$ of scripts with fewest words (mean \#words: $66.8$); S2: $10\%$ of scripts with fewest sentences (mean \#sentences: $3.8$); S3: $10\%$ of scripts with most words (mean \#words: $157.2$); S4: $10\%$ of scripts with most sentences (mean \#sentences: $8.9$). We observed very small performance fluctuations across subsets S1 to S4 (F-Score = $24.5$, $24.5$, $24.7$, $25.0$, respectively); SD-VSum performs slightly better for scripts made of many sentences.

\begin{table}[t]
\caption{Script-driven video summarization: performance comparison (F-Score (\%)) with the CLIP-It method and several variants of the proposed SD-VSum network architecture.}
\label{tab:table1}
\resizebox{\columnwidth}{!}{%
\begin{tabular}{|c|c|c|c|c|}
\hline
          & \begin{tabular}[c]{@{}c@{}}Text\\ representation\end{tabular} & \# Heads & \begin{tabular}[c]{@{}c@{}}Scaling in\\ attention\end{tabular} & F-Score   \\ \hline
SD-VSum   & multiple vectors                                              & 8                                                            & No                                                             & \textbf{24.8} \\
CLIP-It \cite{10.5555/3540261.3541333} & single vector                                                 & 4                                                            & Yes                                                            & 22.8          \\ \hline
Variant 1 & single vector                                                 & 8                                                            & No                                                             & 22.8          \\
Variant 2 & single vector                                                 & 8                                                            & Yes                                                            & 22.7          \\
Variant 3 & single vector                                                 & 4                                                            & Yes                                                            & 22.2          \\
Variant 4 & multiple vectors                                              & 8                                                            & Yes                                                            & 24.2          \\ \hline
\end{tabular}}
\end{table}

Following, since SD-VSum and CLIP-It are similar in terms of network architecture, we tried to spot what causes the observed difference in their performance. For this, we ran an ablation study including the following variants of SD-VSum that resemble the network architecture of CLIP-It, as well as a variant that performs a scaling process when computing the attention weights:
\begin{itemize}
    \item Variant \#1 applies the same single vector representation approach for the user script, with CLIP-It. 
    \item Variant \#2 extends Variant \#1 by containing the same number of attention heads with CLIP-It. 
    \item Variant \#3 extends Variant \#2 by performing a scaling process when computing the attention weights (by dividing with $\sqrt{D}$, where $D$ is the size of the used embeddings), simulating further the language-guided attention mechanism of CLIP-It.
    \item Variant \#4 uses multiple embeddings for the user script and maintains the same number of attention heads ($8$), but performs the aforementioned scaling process when computing the attention weights.
\end{itemize}
Focusing on the lower part of Table \ref{tab:table1}, the performance of Variant \#1 shows that the use of a significantly more condensed representation for the user script leads to a noticeable drop in the summarization performance ($2\%$ lower than SD-VSum). The performance of Variant \#2 indicates the slight effect of the scaling process when using a single vector representation for the user script ($0.1\%$ lower than that of Variant \#1), an effect that is more pronounced when the script is represented using multiple embeddings, as denoted by the performance of Variant \#4 which is $0.6\%$ lower than that of SD-VSum. Finally, the use of fewer attention heads leads to further performance reduction ($0.5\%$ lower than the performance of Variant \#2). These findings document that the use of multiple sentence-level embeddings for text representation (a change compared to CLIP-It) by the integrated cross-modal attention mechanism is critical for addressing the new problem of script-driven summarization, and demonstrate the competency of SD-VSum to deal with scripts describing multiple unrelated actions.

Subsequently, we assessed the capacity of generic video summarization methods to tackle the script-driven video summarization task. For this, we evaluated the performance of SD-VSum and CLIP-It using as input a textual description of the full-length video. Moreover, we ran experiments with a SOTA method for visual-based video summarization (PGL-SUM \cite{9666088}) which estimates frames' importance after modeling their dependence at different levels of granularity using multi-head attention mechanisms. As a note, the computation of the Kendall's $\tau$ and Spearman's $\rho$ rank correlation coefficients for script-driven video summarization is not possible, since the network's output is compared with a binary ground-truth summary. Our comparisons included also the visual-based and multimodal network architectures evaluated by the authors of the VideoXum dataset, using the scores reported in \cite{10334011}. The outcomes of this investigation (see Table \ref{tab:table2}), show that the visual-based approaches are the least effective ones; thus, using a description of the full-length video as auxiliary input information seems to be beneficial for the summarization process. Concerning the multimodal approaches, the performance of SD-VSum is clearly better than that of CLIP-It, and highly competitive to the one of VTSUM-BLIP. Nevertheless, it is lower than the observed performance when SD-VSum is used for script-driven video summarization. These findings indicate the importance of script-driven video summarization technologies, and their capacity to produce video summaries that are better tailored to the needs of different users.

\begin{table}[t]
\caption{Performance comparison between the proposed SD-VSum network architecture for script-driven video summarization and literature methods for visual-based and multimodal video summarization, in terms of F-Score (\%) and Kendall's $\tau$ and Spearman's $\rho$ rank correlation coefficients.}
\label{tab:table2}
%\resizebox{\columnwidth}{!}{%
\begin{tabular}{|cc|c|c|c|}
\hline
\multicolumn{2}{|c|}{}                                                                                                  & F-Score &  $\tau$ & $\rho$ \\ \hline
\multicolumn{1}{|c|}{Script-driven}                   & SD-VSum  & \textbf{24.8}       & N/A         & N/A          \\ \hline
\multicolumn{1}{|c|}{\multirow{3}{*}{\begin{tabular}[c]{@{}c@{}}Multimodal\\ summarization\end{tabular}}}    & SD-VSum  & 23.6        & 0.188         & 0.247          \\

\multicolumn{1}{|c|}{}                                                                                       & VTSUM-BLIP \cite{10334011} & 23.5        & 0.196         & 0.258          \\
\multicolumn{1}{|c|}{}                                                                                       & CLIP-It \cite{10.5555/3540261.3541333} & 22.8        & 0.154         & 0.203          \\ \hline
\multicolumn{1}{|c|}{\multirow{2}{*}{\begin{tabular}[c]{@{}c@{}}Visual-based\\ summarization\end{tabular}}}    & VSUM-BLIP \cite{10334011} & 23.1        & 0.185         & 0.246          \\
\multicolumn{1}{|c|}{}                                                                                       & PGL-SUM \cite{9666088} & 22.0        & 0.153         & 0.203          \\ \hline
\end{tabular}%}
\end{table}

Finally, in terms of computational complexity, we counted the time needed per training epoch and the number of learnable parameters. The proposed SD-VSum network architecture is a bit less complex than CLIP-It, due to the lack of a trainable linear layer that is responsible for producing the single vector representation of the input text. This difference is reflected in Table \ref{tab:table3}, by the slightly reduced time needed per training epoch and the smaller number of trainable parameters of SD-VSum. So, the observed improvements in the summarization performance are gained through a less memory- and computational-demanding training process.

\begin{table}[t]
\caption{Comparison in terms of computational complexity.}
\label{tab:table3}
\begin{tabular}{|c|c|c|}
\hline
\begin{tabular}[c]{@{}c@{}}Network\\ architecture\end{tabular} & \begin{tabular}[c]{@{}c@{}}Training time\\ (min. / epoch)\end{tabular} & \begin{tabular}[c]{@{}c@{}}\# Parameters\\ (in Millions)\end{tabular} \\ \hline
SD-VSum                                                        &   25.25                                                                   &    45.19                                                                   \\ \hline
CLIP-It \cite{10.5555/3540261.3541333}                                                       &    26.17                                                                 &    45.98                                                                   \\ \hline
\end{tabular}
\end{table}

\subsection{Qualitative analysis}

\begin{figure}[t]
\centering
\includegraphics[width=0.92\columnwidth]{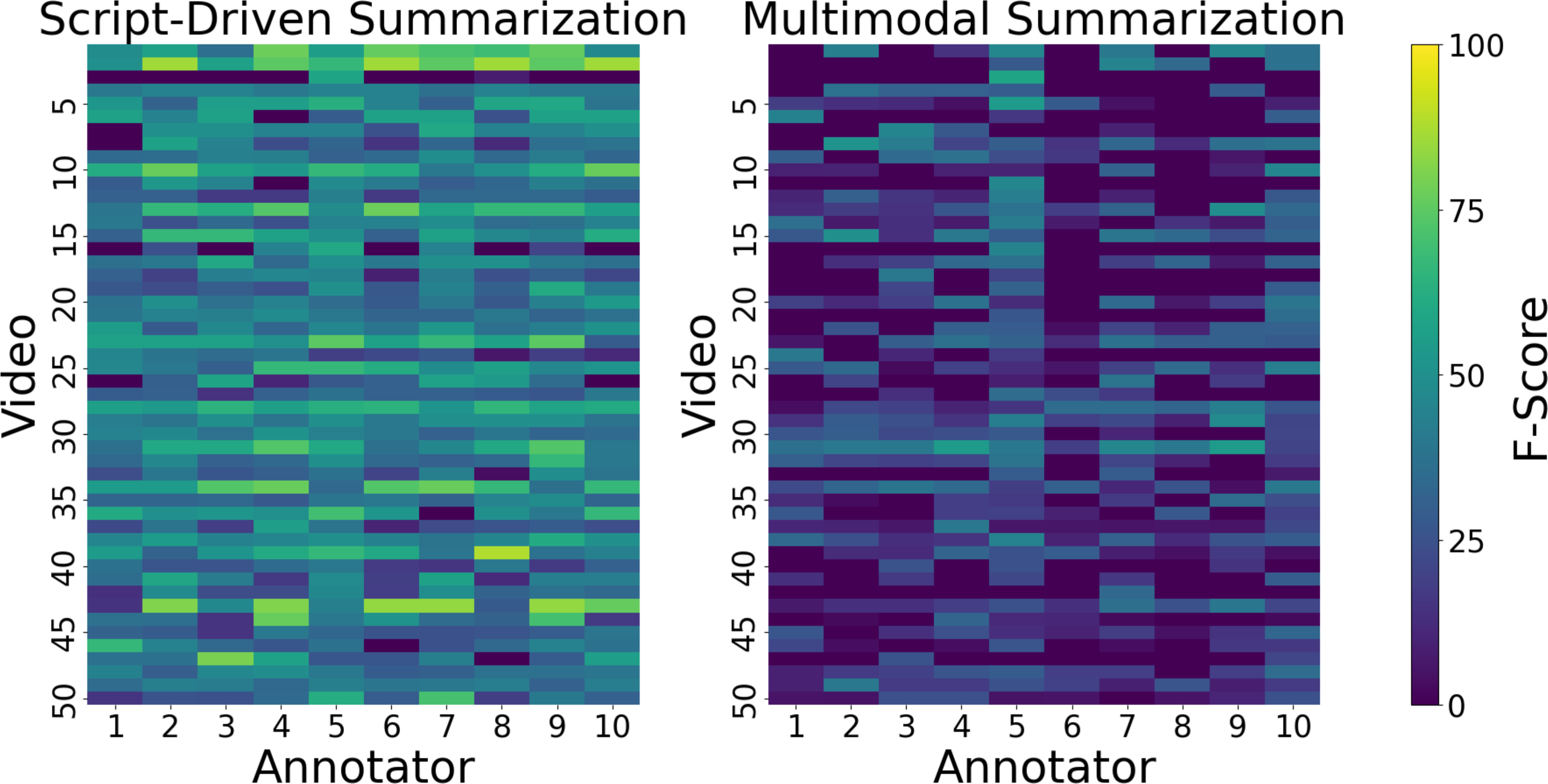}
\caption{The overlap (expressed in F-Score (\%)) between the multiple available ground-truth summaries for $50$ videos of the extended S-VideoXum dataset, and the produced video summaries by SD-VSum when it is used for script-driven (left side) and multimodal video summarization (right side).}
\label{fig:overlap_1}
\end{figure}

Our qualitative analysis aimed to investigate the extent to which the output of script-driven video summarization is more tailored to the needs of different users, compared to summaries produced by multimodal approaches. For this, we compared the overlap (expressed in F-Score (\%)) between the multiple available ground-truth summaries for $50$ videos of the S-VideoXum dataset, and the produced summaries when SD-VSum is used for script-driven and multimodal video summarization, respectively. The pool of videos was defined by choosing the ones showing the major diversity in the computed overlap when comparing the multimodal-based video summary with each different ground-truth summary, and thus were the most challenging ones for a multimodal video summarization approach. The obtained scores, that are illustrated in the form of heatmap values in Fig. \ref{fig:overlap_1}, demonstrate that the output of multimodal video summarization (see the heatmap on the right side) shows the same weak overlap with the summaries of the different human annotators, and thus is poorly adapted to different needs about the content of the summary. On the contrary, the output of script-driven video summarization is clearly more aligned to the preferences of each different annotator, for most of the selected videos. These findings justify once again the motivation behind the development and training of network architectures for script-driven video summarization, and highlight the competency of the proposed SD-VSum network architecture to produce video summaries that are well-tailored to each user's needs about their content.

\section{Conclusions}

In this paper, we introduced a new task that aims to create a video summary of the full-length video by choosing the parts showing the highest relevance with a user-provided script that outlines the content of the desired summary. To support our research, we extended the VideoXum large-scale dataset for video summarization by generating natural language descriptions of the different ground-truth summaries that are available per video. Using the extended S-VideoXum dataset, we trained and evaluated a new network architecture for script-driven video summarization (SD-VSum), that integrates a cross-modal attention mechanism to align and fuse information from the different data modalities. Our quantitative evaluations highlighted the advanced performance of SD-VSum compared to SOTA methods for query-driven and generic (unimodal and multimodal) video summarization from the literature, and pointed out the impact of using coarse-grained representations for the input text. Our qualitative analysis demonstrated the capacity of SD-VSum to incorporate knowledge about the user's needs and produce well-customized video summaries.

\begin{acks}
This work was supported by project MediaPot ($TAE\Delta K$-06196), carried out within the framework of the National Recovery and Resilience Plan Greece 2.0, funded by the European Union – NextGenerationEU, under the call RESEARCH-CREATE-INNOVATE.
\end{acks}

\balance


\begin{thebibliography}{34}

\ifx \showCODEN    \undefined \def \showCODEN     #1{\unskip}     \fi
\ifx \showISBNx    \undefined \def \showISBNx     #1{\unskip}     \fi
\ifx \showISBNxiii \undefined \def \showISBNxiii  #1{\unskip}     \fi
\ifx \showISSN     \undefined \def \showISSN      #1{\unskip}     \fi
\ifx \showLCCN     \undefined \def \showLCCN      #1{\unskip}     \fi
\ifx \shownote     \undefined \def \shownote      #1{#1}          \fi
\ifx \showarticletitle \undefined \def \showarticletitle #1{#1}   \fi
\ifx \showURL      \undefined \def \showURL       {\relax}        \fi
% invisible to TeX
\providecommand\bibfield[2]{#2}
\providecommand\bibinfo[2]{#2}
\providecommand\natexlab[1]{#1}
\providecommand\showeprint[2][]{arXiv:#2}

\bibitem[Apostolidis et~al\mbox{.}(2020)]%
        {10.1145/3394171.3413632}
\bibfield{author}{\bibinfo{person}{Evlampios Apostolidis}, \bibinfo{person}{Eleni Adamantidou}, \bibinfo{person}{Alexandros~I. Metsai}, \bibinfo{person}{Vasileios Mezaris}, {and} \bibinfo{person}{Ioannis Patras}.} \bibinfo{year}{2020}\natexlab{}.
\newblock \showarticletitle{Performance over Random: A Robust Evaluation Protocol for Video Summarization Methods}. In \bibinfo{booktitle}{\emph{Proceedings of the 28th ACM International Conference on Multimedia}} (Seattle, WA, USA) \emph{(\bibinfo{series}{MM '20})}. \bibinfo{publisher}{Association for Computing Machinery}, \bibinfo{address}{New York, NY, USA}, \bibinfo{pages}{1056–1064}.
\newblock
\showISBNx{9781450379885}
\href{https://doi.org/10.1145/3394171.3413632}{doi:\nolinkurl{10.1145/3394171.3413632}}


\bibitem[Apostolidis et~al\mbox{.}(2021a)]%
        {9594911}
\bibfield{author}{\bibinfo{person}{Evlampios Apostolidis}, \bibinfo{person}{Eleni Adamantidou}, \bibinfo{person}{Alexandros~I. Metsai}, \bibinfo{person}{Vasileios Mezaris}, {and} \bibinfo{person}{Ioannis Patras}.} \bibinfo{year}{2021}\natexlab{a}.
\newblock \showarticletitle{Video Summarization Using Deep Neural Networks: A Survey}.
\newblock \bibinfo{journal}{\emph{Proc. IEEE}} \bibinfo{volume}{109}, \bibinfo{number}{11} (\bibinfo{year}{2021}), \bibinfo{pages}{1838--1863}.
\newblock
\href{https://doi.org/10.1109/JPROC.2021.3117472}{doi:\nolinkurl{10.1109/JPROC.2021.3117472}}


\bibitem[Apostolidis et~al\mbox{.}(2021b)]%
        {9666088}
\bibfield{author}{\bibinfo{person}{Evlampios Apostolidis}, \bibinfo{person}{Georgios Balaouras}, \bibinfo{person}{Vasileios Mezaris}, {and} \bibinfo{person}{Ioannis Patras}.} \bibinfo{year}{2021}\natexlab{b}.
\newblock \showarticletitle{Combining Global and Local Attention with Positional Encoding for Video Summarization}. In \bibinfo{booktitle}{\emph{2021 IEEE International Symposium on Multimedia (ISM)}}. \bibinfo{pages}{226--234}.
\newblock


\bibitem[Argaw et~al\mbox{.}(2024)]%
        {10656029}
\bibfield{author}{\bibinfo{person}{Dawit~Mureja Argaw}, \bibinfo{person}{Seunghyun Yoon}, \bibinfo{person}{Fabian~Caba Heilbron}, \bibinfo{person}{Hanieh Deilamsalehy}, \bibinfo{person}{Trung Bui}, \bibinfo{person}{Zhaowen Wang}, \bibinfo{person}{Franck Dernoncourt}, {and} \bibinfo{person}{Joon~Son Chung}.} \bibinfo{year}{2024}\natexlab{}.
\newblock \showarticletitle{{ Scaling Up Video Summarization Pretraining with Large Language Models }}. In \bibinfo{booktitle}{\emph{2024 IEEE/CVF Conference on Computer Vision and Pattern Recognition (CVPR)}}. \bibinfo{publisher}{IEEE Computer Society}, \bibinfo{address}{Los Alamitos, CA, USA}, \bibinfo{pages}{8332--8341}.
\newblock
\href{https://doi.org/10.1109/CVPR52733.2024.00796}{doi:\nolinkurl{10.1109/CVPR52733.2024.00796}}


\bibitem[Cizmeciler et~al\mbox{.}(2022)]%
        {10.1007/s11042-022-12442-w}
\bibfield{author}{\bibinfo{person}{Kemal Cizmeciler}, \bibinfo{person}{Erkut Erdem}, {and} \bibinfo{person}{Aykut Erdem}.} \bibinfo{year}{2022}\natexlab{}.
\newblock \showarticletitle{Leveraging semantic saliency maps for query-specific video summarization}.
\newblock \bibinfo{journal}{\emph{Multimedia Tools Appl.}} \bibinfo{volume}{81}, \bibinfo{number}{12} (\bibinfo{date}{May} \bibinfo{year}{2022}), \bibinfo{pages}{17457–17482}.
\newblock
\showISSN{1380-7501}
\href{https://doi.org/10.1007/s11042-022-12442-w}{doi:\nolinkurl{10.1007/s11042-022-12442-w}}


\bibitem[Gygli et~al\mbox{.}(2014)]%
        {10.1007/978-3-319-10584-0_33}
\bibfield{author}{\bibinfo{person}{Michael Gygli}, \bibinfo{person}{Helmut Grabner}, \bibinfo{person}{Hayko Riemenschneider}, {and} \bibinfo{person}{Luc Van~Gool}.} \bibinfo{year}{2014}\natexlab{}.
\newblock \showarticletitle{Creating Summaries from User Videos}. In \bibinfo{booktitle}{\emph{Computer Vision -- ECCV 2014}}, \bibfield{editor}{\bibinfo{person}{David Fleet}, \bibinfo{person}{Tomas Pajdla}, \bibinfo{person}{Bernt Schiele}, {and} \bibinfo{person}{Tinne Tuytelaars}} (Eds.). \bibinfo{publisher}{Springer International Publishing}, \bibinfo{address}{Cham}, \bibinfo{pages}{505--520}.
\newblock
\showISBNx{978-3-319-10584-0}


\bibitem[He et~al\mbox{.}(2023)]%
        {10204014}
\bibfield{author}{\bibinfo{person}{Bo He}, \bibinfo{person}{Jun Wang}, \bibinfo{person}{Jielin Qiu}, \bibinfo{person}{Trung Bui}, \bibinfo{person}{Abhinav Shrivastava}, {and} \bibinfo{person}{Zhaowen Wang}.} \bibinfo{year}{2023}\natexlab{}.
\newblock \showarticletitle{Align and Attend: Multimodal Summarization with Dual Contrastive Losses}. In \bibinfo{booktitle}{\emph{2023 IEEE/CVF Conference on Computer Vision and Pattern Recognition (CVPR)}}. \bibinfo{pages}{14867--14878}.
\newblock
\href{https://doi.org/10.1109/CVPR52729.2023.01428}{doi:\nolinkurl{10.1109/CVPR52729.2023.01428}}


\bibitem[Hu et~al\mbox{.}(2023)]%
        {Hu2023QuerybasedVS}
\bibfield{author}{\bibinfo{person}{Weifeng Hu}, \bibinfo{person}{Y. Zhang}, \bibinfo{person}{Yujun Li}, \bibinfo{person}{Jia Zhao}, \bibinfo{person}{Xifeng Hu}, \bibinfo{person}{Yan Cui}, {and} \bibinfo{person}{Xuejing Wang}.} \bibinfo{year}{2023}\natexlab{}.
\newblock \showarticletitle{Query-based video summarization with multi-label classification network}.
\newblock \bibinfo{journal}{\emph{Multimedia Tools and Applications}}  \bibinfo{volume}{82} (\bibinfo{year}{2023}), \bibinfo{pages}{37529--37549}.
\newblock
\urldef\tempurl%
\url{https://doi.org/10.1007/s11042-023-15126-1}
\showURL{%
\tempurl}


\bibitem[Huang et~al\mbox{.}(2023)]%
        {10222138}
\bibfield{author}{\bibinfo{person}{Jia-Hong Huang}, \bibinfo{person}{Luka Murn}, \bibinfo{person}{Marta Mrak}, {and} \bibinfo{person}{Marcel Worring}.} \bibinfo{year}{2023}\natexlab{}.
\newblock \showarticletitle{Query-Based Video Summarization with Pseudo Label Supervision}. In \bibinfo{booktitle}{\emph{2023 IEEE International Conference on Image Processing (ICIP)}}. \bibinfo{pages}{1430--1434}.
\newblock
\href{https://doi.org/10.1109/ICIP49359.2023.10222138}{doi:\nolinkurl{10.1109/ICIP49359.2023.10222138}}


\bibitem[Huang and Worring(2020)]%
        {10.1145/3372278.3390695}
\bibfield{author}{\bibinfo{person}{Jia-Hong Huang} {and} \bibinfo{person}{Marcel Worring}.} \bibinfo{year}{2020}\natexlab{}.
\newblock \showarticletitle{Query-controllable Video Summarization}. In \bibinfo{booktitle}{\emph{Proceedings of the 2020 International Conference on Multimedia Retrieval}} (Dublin, Ireland) \emph{(\bibinfo{series}{ICMR '20})}. \bibinfo{publisher}{Association for Computing Machinery}, \bibinfo{address}{New York, NY, USA}, \bibinfo{pages}{242–250}.
\newblock
\showISBNx{9781450370875}
\href{https://doi.org/10.1145/3372278.3390695}{doi:\nolinkurl{10.1145/3372278.3390695}}


\bibitem[Jiang and Han(2019)]%
        {10.1145/3323873.3325040}
\bibfield{author}{\bibinfo{person}{Pin Jiang} {and} \bibinfo{person}{Yahong Han}.} \bibinfo{year}{2019}\natexlab{}.
\newblock \showarticletitle{Hierarchical Variational Network for User-Diversified \& Query-Focused Video Summarization}. In \bibinfo{booktitle}{\emph{Proceedings of the 2019 on International Conference on Multimedia Retrieval}} (Ottawa ON, Canada) \emph{(\bibinfo{series}{ICMR '19})}. \bibinfo{publisher}{Association for Computing Machinery}, \bibinfo{address}{New York, NY, USA}, \bibinfo{pages}{202–206}.
\newblock
\showISBNx{9781450367653}
\href{https://doi.org/10.1145/3323873.3325040}{doi:\nolinkurl{10.1145/3323873.3325040}}


\bibitem[Kadam and Deshpande(2024)]%
        {Kadam2024}
\bibfield{author}{\bibinfo{person}{Bhakti~D. Kadam} {and} \bibinfo{person}{Ashwini~M. Deshpande}.} \bibinfo{year}{2024}\natexlab{}.
\newblock \showarticletitle{Query-attentive video summarization: A comprehensive review}.
\newblock \bibinfo{journal}{\emph{Multimedia Tools and Applications}} (\bibinfo{date}{06 Aug} \bibinfo{year}{2024}).
\newblock
\showISSN{1573-7721}
\href{https://doi.org/10.1007/s11042-024-19977-0}{doi:\nolinkurl{10.1007/s11042-024-19977-0}}


\bibitem[Kendall(1945)]%
        {kendall1945treatment}
\bibfield{author}{\bibinfo{person}{Maurice~G Kendall}.} \bibinfo{year}{1945}\natexlab{}.
\newblock \showarticletitle{{The treatment of ties in ranking problems}}.
\newblock \bibinfo{journal}{\emph{Biometrika}} \bibinfo{volume}{33}, \bibinfo{number}{3} (\bibinfo{year}{1945}), \bibinfo{pages}{239--251}.
\newblock


\bibitem[Kokoska and Zwillinger(2000)]%
        {kokoska2000crc}
\bibfield{author}{\bibinfo{person}{Stephen Kokoska} {and} \bibinfo{person}{Daniel Zwillinger}.} \bibinfo{year}{2000}\natexlab{}.
\newblock \bibinfo{booktitle}{\emph{{CRC standard probability and statistics tables and formulae}}}.
\newblock \bibinfo{publisher}{Crc Press}.
\newblock


\bibitem[Krishna et~al\mbox{.}(2017)]%
        {8237345}
\bibfield{author}{\bibinfo{person}{Ranjay Krishna}, \bibinfo{person}{Kenji Hata}, \bibinfo{person}{Frederic Ren}, \bibinfo{person}{Li Fei-Fei}, {and} \bibinfo{person}{Juan~Carlos Niebles}.} \bibinfo{year}{2017}\natexlab{}.
\newblock \showarticletitle{Dense-Captioning Events in Videos}. In \bibinfo{booktitle}{\emph{2017 IEEE International Conference on Computer Vision (ICCV)}}. \bibinfo{pages}{706--715}.
\newblock
\href{https://doi.org/10.1109/ICCV.2017.83}{doi:\nolinkurl{10.1109/ICCV.2017.83}}


\bibitem[Lan et~al\mbox{.}(2023)]%
        {10.1145/3532626}
\bibfield{author}{\bibinfo{person}{Xiaohan Lan}, \bibinfo{person}{Yitian Yuan}, \bibinfo{person}{Xin Wang}, \bibinfo{person}{Zhi Wang}, {and} \bibinfo{person}{Wenwu Zhu}.} \bibinfo{year}{2023}\natexlab{}.
\newblock \showarticletitle{A Survey on Temporal Sentence Grounding in Videos}.
\newblock \bibinfo{journal}{\emph{ACM Trans. Multimedia Comput. Commun. Appl.}} \bibinfo{volume}{19}, \bibinfo{number}{2}, Article \bibinfo{articleno}{51} (\bibinfo{date}{Feb.} \bibinfo{year}{2023}), \bibinfo{numpages}{33}~pages.
\newblock
\showISSN{1551-6857}
\href{https://doi.org/10.1145/3532626}{doi:\nolinkurl{10.1145/3532626}}


\bibitem[Lee et~al\mbox{.}(2012)]%
        {6247820}
\bibfield{author}{\bibinfo{person}{Yong~Jae Lee}, \bibinfo{person}{Joydeep Ghosh}, {and} \bibinfo{person}{Kristen Grauman}.} \bibinfo{year}{2012}\natexlab{}.
\newblock \showarticletitle{Discovering important people and objects for egocentric video summarization}. In \bibinfo{booktitle}{\emph{2012 IEEE Conference on Computer Vision and Pattern Recognition}}. \bibinfo{pages}{1346--1353}.
\newblock
\href{https://doi.org/10.1109/CVPR.2012.6247820}{doi:\nolinkurl{10.1109/CVPR.2012.6247820}}


\bibitem[Li et~al\mbox{.}(2024)]%
        {li2024llavanext-strong}
\bibfield{author}{\bibinfo{person}{Bo Li}, \bibinfo{person}{Kaichen Zhang}, \bibinfo{person}{Hao Zhang}, \bibinfo{person}{Dong Guo}, \bibinfo{person}{Renrui Zhang}, \bibinfo{person}{Feng Li}, \bibinfo{person}{Yuanhan Zhang}, \bibinfo{person}{Ziwei Liu}, {and} \bibinfo{person}{Chunyuan Li}.} \bibinfo{year}{2024}\natexlab{}.
\newblock \bibinfo{title}{LLaVA-NeXT: Stronger LLMs Supercharge Multimodal Capabilities in the Wild}.
\newblock
\urldef\tempurl%
\url{https://llava-vl.github.io/blog/2024-05-10-llava-next-stronger-llms/}
\showURL{%
\tempurl}


\bibitem[Lin(2004)]%
        {lin-2004-rouge}
\bibfield{author}{\bibinfo{person}{Chin-Yew Lin}.} \bibinfo{year}{2004}\natexlab{}.
\newblock \showarticletitle{{ROUGE}: A Package for Automatic Evaluation of Summaries}. In \bibinfo{booktitle}{\emph{Text Summarization Branches Out}}. \bibinfo{publisher}{Association for Computational Linguistics}, \bibinfo{address}{Barcelona, Spain}, \bibinfo{pages}{74--81}.
\newblock
\urldef\tempurl%
\url{https://aclanthology.org/W04-1013/}
\showURL{%
\tempurl}


\bibitem[Lin et~al\mbox{.}(2024)]%
        {10334011}
\bibfield{author}{\bibinfo{person}{Jingyang Lin}, \bibinfo{person}{Hang Hua}, \bibinfo{person}{Ming Chen}, \bibinfo{person}{Yikang Li}, \bibinfo{person}{Jenhao Hsiao}, \bibinfo{person}{Chiuman Ho}, {and} \bibinfo{person}{Jiebo Luo}.} \bibinfo{year}{2024}\natexlab{}.
\newblock \showarticletitle{VideoXum: Cross-Modal Visual and Textural Summarization of Videos}.
\newblock \bibinfo{journal}{\emph{IEEE Transactions on Multimedia}}  \bibinfo{volume}{26} (\bibinfo{year}{2024}), \bibinfo{pages}{5548--5560}.
\newblock
\href{https://doi.org/10.1109/TMM.2023.3335875}{doi:\nolinkurl{10.1109/TMM.2023.3335875}}


\bibitem[Narasimhan et~al\mbox{.}(2022)]%
        {10.1007/978-3-031-19830-4_31}
\bibfield{author}{\bibinfo{person}{Medhini Narasimhan}, \bibinfo{person}{Arsha Nagrani}, \bibinfo{person}{Chen Sun}, \bibinfo{person}{Michael Rubinstein}, \bibinfo{person}{Trevor Darrell}, \bibinfo{person}{Anna Rohrbach}, {and} \bibinfo{person}{Cordelia Schmid}.} \bibinfo{year}{2022}\natexlab{}.
\newblock \showarticletitle{TL;DW? Summarizing Instructional Videos with Task Relevance and Cross-Modal Saliency}. In \bibinfo{booktitle}{\emph{Computer Vision -- ECCV 2022}}, \bibfield{editor}{\bibinfo{person}{Shai Avidan}, \bibinfo{person}{Gabriel Brostow}, \bibinfo{person}{Moustapha Ciss{\'e}}, \bibinfo{person}{Giovanni~Maria Farinella}, {and} \bibinfo{person}{Tal Hassner}} (Eds.). \bibinfo{publisher}{Springer Nature Switzerland}, \bibinfo{address}{Cham}, \bibinfo{pages}{540--557}.
\newblock
\showISBNx{978-3-031-19830-4}


\bibitem[Narasimhan et~al\mbox{.}(2021)]%
        {10.5555/3540261.3541333}
\bibfield{author}{\bibinfo{person}{Medhini Narasimhan}, \bibinfo{person}{Anna Rohrbach}, {and} \bibinfo{person}{Trevor Darrell}.} \bibinfo{year}{2021}\natexlab{}.
\newblock \showarticletitle{CLIP-It! language-guided video summarization}. In \bibinfo{booktitle}{\emph{Proceedings of the 35th International Conference on Neural Information Processing Systems}} \emph{(\bibinfo{series}{NIPS '21})}. \bibinfo{publisher}{Curran Associates Inc.}, \bibinfo{address}{Red Hook, NY, USA}, Article \bibinfo{articleno}{1072}, \bibinfo{numpages}{13}~pages.
\newblock
\showISBNx{9781713845393}


\bibitem[Otani et~al\mbox{.}(2019)]%
        {8954229}
\bibfield{author}{\bibinfo{person}{Mayu Otani}, \bibinfo{person}{Yuta Nakashima}, \bibinfo{person}{Esa Rahtu}, {and} \bibinfo{person}{Janne Heikkilä}.} \bibinfo{year}{2019}\natexlab{}.
\newblock \showarticletitle{Rethinking the Evaluation of Video Summaries}. In \bibinfo{booktitle}{\emph{2019 IEEE/CVF Conference on Computer Vision and Pattern Recognition (CVPR)}}. \bibinfo{pages}{7588--7596}.
\newblock
\href{https://doi.org/10.1109/CVPR.2019.00778}{doi:\nolinkurl{10.1109/CVPR.2019.00778}}


\bibitem[Sharghi et~al\mbox{.}(2016)]%
        {10.1007/978-3-319-46484-8_1}
\bibfield{author}{\bibinfo{person}{Aidean Sharghi}, \bibinfo{person}{Boqing Gong}, {and} \bibinfo{person}{Mubarak Shah}.} \bibinfo{year}{2016}\natexlab{}.
\newblock \showarticletitle{Query-Focused Extractive Video Summarization}. In \bibinfo{booktitle}{\emph{Computer Vision -- ECCV 2016}}, \bibfield{editor}{\bibinfo{person}{Bastian Leibe}, \bibinfo{person}{Jiri Matas}, \bibinfo{person}{Nicu Sebe}, {and} \bibinfo{person}{Max Welling}} (Eds.). \bibinfo{publisher}{Springer International Publishing}, \bibinfo{address}{Cham}, \bibinfo{pages}{3--19}.
\newblock
\showISBNx{978-3-319-46484-8}


\bibitem[Sharghi et~al\mbox{.}(2017)]%
        {8099712}
\bibfield{author}{\bibinfo{person}{Aidean Sharghi}, \bibinfo{person}{Jacob~S. Laurel}, {and} \bibinfo{person}{Boqing Gong}.} \bibinfo{year}{2017}\natexlab{}.
\newblock \showarticletitle{Query-Focused Video Summarization: Dataset, Evaluation, and a Memory Network Based Approach}. In \bibinfo{booktitle}{\emph{2017 IEEE Conference on Computer Vision and Pattern Recognition (CVPR)}}. \bibinfo{pages}{2127--2136}.
\newblock
\href{https://doi.org/10.1109/CVPR.2017.229}{doi:\nolinkurl{10.1109/CVPR.2017.229}}


\bibitem[Song et~al\mbox{.}(2015)]%
        {7299154}
\bibfield{author}{\bibinfo{person}{Yale Song}, \bibinfo{person}{Jordi Vallmitjana}, \bibinfo{person}{Amanda Stent}, {and} \bibinfo{person}{Alejandro Jaimes}.} \bibinfo{year}{2015}\natexlab{}.
\newblock \showarticletitle{TVSum: Summarizing web videos using titles}. In \bibinfo{booktitle}{\emph{2015 IEEE Conference on Computer Vision and Pattern Recognition (CVPR)}}. \bibinfo{pages}{5179--5187}.
\newblock
\href{https://doi.org/10.1109/CVPR.2015.7299154}{doi:\nolinkurl{10.1109/CVPR.2015.7299154}}


\bibitem[Vasudevan et~al\mbox{.}(2017)]%
        {10.1145/3123266.3123297}
\bibfield{author}{\bibinfo{person}{Arun~Balajee Vasudevan}, \bibinfo{person}{Michael Gygli}, \bibinfo{person}{Anna Volokitin}, {and} \bibinfo{person}{Luc Van~Gool}.} \bibinfo{year}{2017}\natexlab{}.
\newblock \showarticletitle{Query-adaptive Video Summarization via Quality-aware Relevance Estimation}. In \bibinfo{booktitle}{\emph{Proceedings of the 25th ACM International Conference on Multimedia}} (Mountain View, California, USA) \emph{(\bibinfo{series}{MM '17})}. \bibinfo{publisher}{Association for Computing Machinery}, \bibinfo{address}{New York, NY, USA}, \bibinfo{pages}{582–590}.
\newblock
\showISBNx{9781450349062}
\href{https://doi.org/10.1145/3123266.3123297}{doi:\nolinkurl{10.1145/3123266.3123297}}


\bibitem[Vaswani et~al\mbox{.}(2017)]%
        {10.5555/3295222.3295349}
\bibfield{author}{\bibinfo{person}{Ashish Vaswani}, \bibinfo{person}{Noam Shazeer}, \bibinfo{person}{Niki Parmar}, \bibinfo{person}{Jakob Uszkoreit}, \bibinfo{person}{Llion Jones}, \bibinfo{person}{Aidan~N. Gomez}, \bibinfo{person}{\L{}ukasz Kaiser}, {and} \bibinfo{person}{Illia Polosukhin}.} \bibinfo{year}{2017}\natexlab{}.
\newblock \showarticletitle{Attention is all you need}. In \bibinfo{booktitle}{\emph{Proceedings of the 31st International Conference on Neural Information Processing Systems}} (Long Beach, California, USA) \emph{(\bibinfo{series}{NIPS'17})}. \bibinfo{publisher}{Curran Associates Inc.}, \bibinfo{address}{Red Hook, NY, USA}, \bibinfo{pages}{6000–6010}.
\newblock
\showISBNx{9781510860964}


\bibitem[Wei et~al\mbox{.}(2018)]%
        {10.5555/3504035.3504062}
\bibfield{author}{\bibinfo{person}{Huawei Wei}, \bibinfo{person}{Bingbing Ni}, \bibinfo{person}{Yichao Yan}, \bibinfo{person}{Huanyu Yu}, {and} \bibinfo{person}{Xiaokang Yang}.} \bibinfo{year}{2018}\natexlab{}.
\newblock \showarticletitle{Video summarization via semantic attended networks}. In \bibinfo{booktitle}{\emph{Proceedings of the Thirty-Second AAAI Conference on Artificial Intelligence and Thirtieth Innovative Applications of Artificial Intelligence Conference and Eighth AAAI Symposium on Educational Advances in Artificial Intelligence}} (New Orleans, Louisiana, USA) \emph{(\bibinfo{series}{AAAI'18/IAAI'18/EAAI'18})}. \bibinfo{publisher}{AAAI Press}, Article \bibinfo{articleno}{27}, \bibinfo{numpages}{8}~pages.
\newblock
\showISBNx{978-1-57735-800-8}


\bibitem[Xiao et~al\mbox{.}(2020a)]%
        {9063637}
\bibfield{author}{\bibinfo{person}{Shuwen Xiao}, \bibinfo{person}{Zhou Zhao}, \bibinfo{person}{Zijian Zhang}, \bibinfo{person}{Ziyu Guan}, {and} \bibinfo{person}{Deng Cai}.} \bibinfo{year}{2020}\natexlab{a}.
\newblock \showarticletitle{Query-Biased Self-Attentive Network for Query-Focused Video Summarization}.
\newblock \bibinfo{journal}{\emph{IEEE Transactions on Image Processing}}  \bibinfo{volume}{29} (\bibinfo{year}{2020}), \bibinfo{pages}{5889--5899}.
\newblock
\href{https://doi.org/10.1109/TIP.2020.2985868}{doi:\nolinkurl{10.1109/TIP.2020.2985868}}


\bibitem[Xiao et~al\mbox{.}(2020b)]%
        {Xiao_Zhao_Zhang_Yan_Yang_2020}
\bibfield{author}{\bibinfo{person}{Shuwen Xiao}, \bibinfo{person}{Zhou Zhao}, \bibinfo{person}{Zijian Zhang}, \bibinfo{person}{Xiaohui Yan}, {and} \bibinfo{person}{Min Yang}.} \bibinfo{year}{2020}\natexlab{b}.
\newblock \showarticletitle{Convolutional Hierarchical Attention Network for Query-Focused Video Summarization}.
\newblock \bibinfo{journal}{\emph{Proceedings of the AAAI Conference on Artificial Intelligence}} \bibinfo{volume}{34}, \bibinfo{number}{07} (\bibinfo{date}{Apr.} \bibinfo{year}{2020}), \bibinfo{pages}{12426--12433}.
\newblock
\href{https://doi.org/10.1609/aaai.v34i07.6929}{doi:\nolinkurl{10.1609/aaai.v34i07.6929}}


\bibitem[Yeung et~al\mbox{.}(2014)]%
        {Yeung2014VideoSETVS}
\bibfield{author}{\bibinfo{person}{Serena Yeung}, \bibinfo{person}{Alireza Fathi}, {and} \bibinfo{person}{Li Fei-Fei}.} \bibinfo{year}{2014}\natexlab{}.
\newblock \showarticletitle{VideoSET: Video Summary Evaluation through Text}.
\newblock \bibinfo{journal}{\emph{ArXiv}}  \bibinfo{volume}{abs/1406.5824} (\bibinfo{year}{2014}).
\newblock
\urldef\tempurl%
\url{https://api.semanticscholar.org/CorpusID:9338736}
\showURL{%
\tempurl}


\bibitem[Zhang et~al\mbox{.}(2018)]%
        {Zhang2018QCVS}
\bibfield{author}{\bibinfo{person}{Yujia Zhang}, \bibinfo{person}{Michael~C. Kampffmeyer}, \bibinfo{person}{Xiaodan Liang}, \bibinfo{person}{Min Tan}, {and} \bibinfo{person}{Eric~P. Xing}.} \bibinfo{year}{2018}\natexlab{}.
\newblock \showarticletitle{{Query-Conditioned Three-Player Adversarial Network for Video Summarization}}. In \bibinfo{booktitle}{\emph{Proceedings of the 2018 British Machine Vision Conf. (BMVC)}}.
\newblock


\bibitem[Zhong et~al\mbox{.}(2022)]%
        {10.1145/3477538}
\bibfield{author}{\bibinfo{person}{Sheng-Hua Zhong}, \bibinfo{person}{Jingxu Lin}, \bibinfo{person}{Jianglin Lu}, \bibinfo{person}{Ahmed Fares}, {and} \bibinfo{person}{Tongwei Ren}.} \bibinfo{year}{2022}\natexlab{}.
\newblock \showarticletitle{Deep Semantic and Attentive Network for Unsupervised Video Summarization}.
\newblock \bibinfo{journal}{\emph{ACM Trans. Multimedia Comput. Commun. Appl.}} \bibinfo{volume}{18}, \bibinfo{number}{2}, Article \bibinfo{articleno}{55} (\bibinfo{date}{Feb.} \bibinfo{year}{2022}), \bibinfo{numpages}{21}~pages.
\newblock
\showISSN{1551-6857}
\href{https://doi.org/10.1145/3477538}{doi:\nolinkurl{10.1145/3477538}}


\end{thebibliography}
\end{document}